\documentclass{article}

\usepackage{arxiv}

\usepackage[utf8]{inputenc} 
\usepackage[T1]{fontenc}    

\usepackage{url}            
\usepackage{booktabs}       
\usepackage{amsfonts}       
\usepackage{nicefrac}       
\usepackage{microtype}      
\usepackage{lipsum}
\usepackage{amssymb}
\usepackage{latexsym}

\usepackage{url}
\usepackage{xcolor}
\usepackage{hyperref}

\usepackage{amsmath}
\usepackage{graphicx}
\usepackage{cleveref}
\usepackage[export]{adjustbox}
\usepackage{subfig}
\usepackage{mwe}
\usepackage{graphicx}
\graphicspath{ {./images/} }

\title{Sperm Detection and Tracking in Phase-Contrast Microscopy Image Sequences Using Deep Learning and modified CSR-DCF}

\author{
 Mohammad reza Mohammadi \\
  School of Computer Engineering\\
  Iran University of Science and Technology, Iran\\
  \texttt{mrmohammadi@iust.ac.ir} \\
   \And
 Mohammad Rahimzadeh \\
  School of Computer Engineering\\
  Iran University of Science and Technology, Iran\\
  \texttt{mh\_rahimzadeh@elec.iust.ac.ir} \\
  \And
 Abolfazl Attar \\
  Department of Electrical Engineering\\
  Sharif University of Technology, Iran\\
  \texttt{attar.abolfazl@ee.sharif.edu} \\
}

\begin{document}
\maketitle
\begin{abstract}
Nowadays, computer-aided sperm analysis (CASA) systems have made a big leap in extracting the characteristics of spermatozoa for studies or measuring human fertility.
The first step in sperm characteristics analysis is sperm detection in the frames of the video sample.
In this article, we used RetinaNet, a deep fully convolutional neural network as the object detector.
Sperms are small objects with few attributes, that makes the detection more difficult in high-density samples and especially when there are other particles in semen, which could be like sperm heads.
One of the main attributes of sperms is their movement, but this attribute cannot be extracted when only one frame would be fed to the network.
To improve the performance of the sperm detection network, we concatenated some consecutive frames to use as the input of the network. With this method, the motility attribute has also been extracted, and then with the help of the deep convolutional network, we have achieved high accuracy in sperm detection.
The second step is tracking the sperms, for extracting the motility parameters that are essential for indicating fertility and other studies on sperms.
In the tracking phase, we modify the CSR-DCF algorithm.
This method also has shown excellent results in sperm tracking even in high-density sperm samples, occlusions, sperm colliding, and when sperms exit from a frame and re-enter in the next frames.
The average precision of the detection phase is 99.1\%, and the F1 score of the tracking method evaluation is 96.61\%.
These results can be a great help in studies investigating sperm behavior and analyzing fertility possibility.
\end{abstract}



\section{Introduction}
\label{1}

Scientists have reported that infertility has become a severe problem for couples.
Based on statistics, almost 15\% of couples suffer from infertility problems all over the world \cite{Gussi2005}.
As reported amount couples with infertility problems in the United States, almost 35-40\% of the problems are caused by male partners, and almost 35\% caused by female partners, 20\% are traced to a problem in both partners and 10\% because of unknown reasons \cite{world1999laboratory}.

The ability of fertility in men depends on sperms concentration (existence of enough numbers of sperms in a specific amount of semen), the direction of sperms motility, and their morphology (size and shape of the sperms' head and tail) \cite{el2003human,menkveld2001semen}.
Based on these fertility factors, motion analysis of sperms is very important for determining male fertility. As the detection and tracking of sperms be performed more accurately, it results in a more accurate diagnosis of infertility problems. 
The most common way of analyzing the sperms is through an expert by observing the sperms via microscope and reporting their motion quality, numbers, and morphology, which is difficult \cite{amann2004andrology}.

Besides the manual way, computer-aided sperm analysis (CASA) systems also have been used for sperm analysis.
CASA systems have been improved very much in the past decades and now are performing faster and more accurate than manually observation \cite{zinaman1996evaluation,mortimer2015future}.
CASA systems use different algorithms to obtain specifications from images or video samples of sperms, which some of these attributes are numbers of sperms, morphology, and especially motility parameters \cite{shi2006computer}.
Most of the prior works of CASA are based on the classic image processing and machine learning algorithms \cite{arasteh2018multi}.
In the past years, deep learning has been state of the art in many computer vision domains.
In this paper, we have used deep learning to improve the accuracy of sperm detection.

Deep learning is a branch of machine learning and is implemented on deep neural networks \cite{schmidhuber2015deep}.
It is capable of automatically extracting high-dimensional features from the input raw data \cite{lecun2015deep}.
Nowadays, deep learning is utilized in many domains of science, business, and government, like reconstructing brain circuits \cite{helmstaedter2013connectomic}, predicting the activity of potential drug molecules \cite{ma2015deep}, and predicting the effects of mutations in non-coding DNA on gene expression and disease \cite{xiong2015human}.
With the advent of convolutional neural networks (CNN), image processing speed and accuracy have been improved a lot \cite{oquab2014learning}.
Some of the CNN-based object detectors are R-CNN \cite{ren2015faster}, SSD \cite{liu2016ssd}, YOLO \cite{redmon2017yolo9000} and RetinaNet \cite{lin2018focal}.

    
Sperm detection is the first step of automatic sperm tracking.
In this paper, we use RetinaNet \cite{lin2018focal}, a deep fully convolutional neural network for sperm detection.
Sperm attributes are few, and this makes the work of detector more difficult.
Our novelty for the detection part is to introduce a new method for training and testing the deep neural network when our data is sequences of images, like videos, and our objects are motile.
RetinaNet and other object detectors, firstly extract input features of input data, then those features will be used for object detection.
This novelty helps the network to extract motility attributes plus other attributes, and so, outputs better results.
For implementing this method, instead of giving one frame of video to the network as input, we fed the concatenation of several consecutive frames of a video to the network.
By doing this, the network would be able to identify motion attributes too, so it learns better about sperms and performs the detection more accurately.

For the tracking part, we introduced a new method that uses both detected sperms in the video frames and tracking algorithm.
We named our tracker modified CSR-DCF.
The modified CSR-DCF is a multi-object tracker that uses the CSR-DCF tracker algorithm \cite{lukezic_vojir_zajc_matas_kristan_2017} as its core.
The CSR-DCF is originally a single object tracker, but in our development, we modified it to become a multi-object tracker.
The modified CSR-DCF performs some different algorithms to utilize the detected sperms while tracking, also finds the wrong tracked, and non-tracked sperms and corrects them.
After that, in case of the existence of wrong detected and non-detected sperms that can cause false tracks, the tracker runs an algorithm to correct these as much as possible.
Our developed modified CSR-DCF algorithm is a robust multi-sperm tracker that works accurately, even when sperms collide or cross each other pass.

The rest of this paper is organized as follows.
A review of the existing methods for sperm detection and tracking are presented in Section \ref{2}.
Our proposed approach is presented in Section \ref{3}.
The experimental results are reported in Section \ref{4}, and finally the paper is concluded in Section \ref{5} and in Section \ref{6}, we presented the used code of this paper.

\section{Related works}
\label{2}
Most of the prior works on sperm tracking in the past decades were done based on single-sperm tracking, which does not work well while sperms collide with each other \cite{urbano2016automatic}. Although different studies have been done on multi-sperm tracking in recent years, many of them have been evaluated on a small dataset.\cite{beya2015automatic}

Some of the early studies about sperms tracking that took place in the '80s and '90s are \cite{katz1985real,groenewald1991preprocessing}, which both are single-cell trackers. In a study \cite{shi2006computer}, another single-sperm tracking algorithm is proposed that tracks the sperms by creating a region around them and uses a modified four-class threshold and post-collision analysis would be performed to determine tracked sperms in the images. This method also uses a speed-check feature to track sperms when they are near to other sperms or particles. Research \cite{zhang2018robotic}, aims to develop a robotic system for immobilization of motile sperm (in order to clinical injection) to track a single sperm's head.

\cite{hesar2018multiple} presented another method that used a modified Gaussian mixture probability hypothesis density (GM-PHD) filter for tracking multi sperms simultaneously.
In \cite{sorensen2008multi}, the Laplacian of Gaussian operator was used to detect the head of sperms.
Then, a combination of particle and Kalman filter applied for tracking the sperm's movement.
Another method based on the radar-tracking algorithm \cite{urbano2016automatic} was introduced for automatic detection and multi-sperm tracking simultaneously.
The detection task was done in a sequence of noise reduction by applying a Gaussian filter.
Then highlights sperm's head by applying Laplacian of Gaussian/Mexican Hat filter and Otsu's method for selecting threshold.
After that, it removes objects less than 5 pixels (less probable to be sperm).
A joint probabilistic data association filter performed the tracking task.
It claims that the tracker works well in sperm occlusion.
After tracking the motility, this method was evaluated on only two videos (from two persons) parameters were also extracted for analysis and contained a totally of 717 sperms, while we used a dataset of 36 different persons and containing 1628 sperms.

\cite{beya2015automatic} proposed another method for automatically sperm detection and tracking.
The detection performs in the first frame of video by applying a bag-of-words method and SVM classification, then by using the mean shift method the detected sperms would be tracked in other frames.
Although this study reached a good result, its disadvantage is using a small dataset.
\cite{kheirkhah2018modified} used a non-linear preprocessing and histogram-based thresholding algorithm for sperm detection and an adaptive distance scheme (AWAS algorithm) for the tracking Section.
This method does not work well in medium or high-density samples.
In \cite{Kheirkhah2019}, a new algorithm, Hybrid-Kittler, based on the combination of Kittler and modified Kittler method, is proposed for sperm segmentation and another method for removing fixed sperms to improve tracking speed.
Then, for tracking the detected sperms, the modified adoptive windows average speed (AWAS) algorithm was applied.
After tracking, another method was used to detect the lost sperms and assignment to the tracks.
\cite{arasteh2018multi} introduced a new hybrid dynamic Bayesian network (HDBN) model for multi-target tracking that was evaluated on 1659 manually extracted dataset and achieved fair results for tracking stage, but segmentation results were not so accurate.

The study \cite{riordon2019deep} used VGG16, which is a deep convolutional neural network, for classifying sperm shape on the World Health Organization (WHO) categories.
\cite{javadi2019novel} also used deep learning for analyzing the sperm morphology and detects sperms morphology problems.
The dataset used in this research consisted of 1540 sperms.
In another study \cite{hicks2019machine}, machine learning algorithms like linear regression and other methods based on convolutional neural networks were used to indicate sperms motility, and this study was tested on a dataset consisting of 85 videos.
\cite{bermant2019deep} used convolutional neural networks to Detect and Classify Sperm Whale Bioacoustics.

\section{Proposed Approach}
\label{3}
\begin{figure*}[!t]
	\centering
	\includegraphics[width=1.0\linewidth]{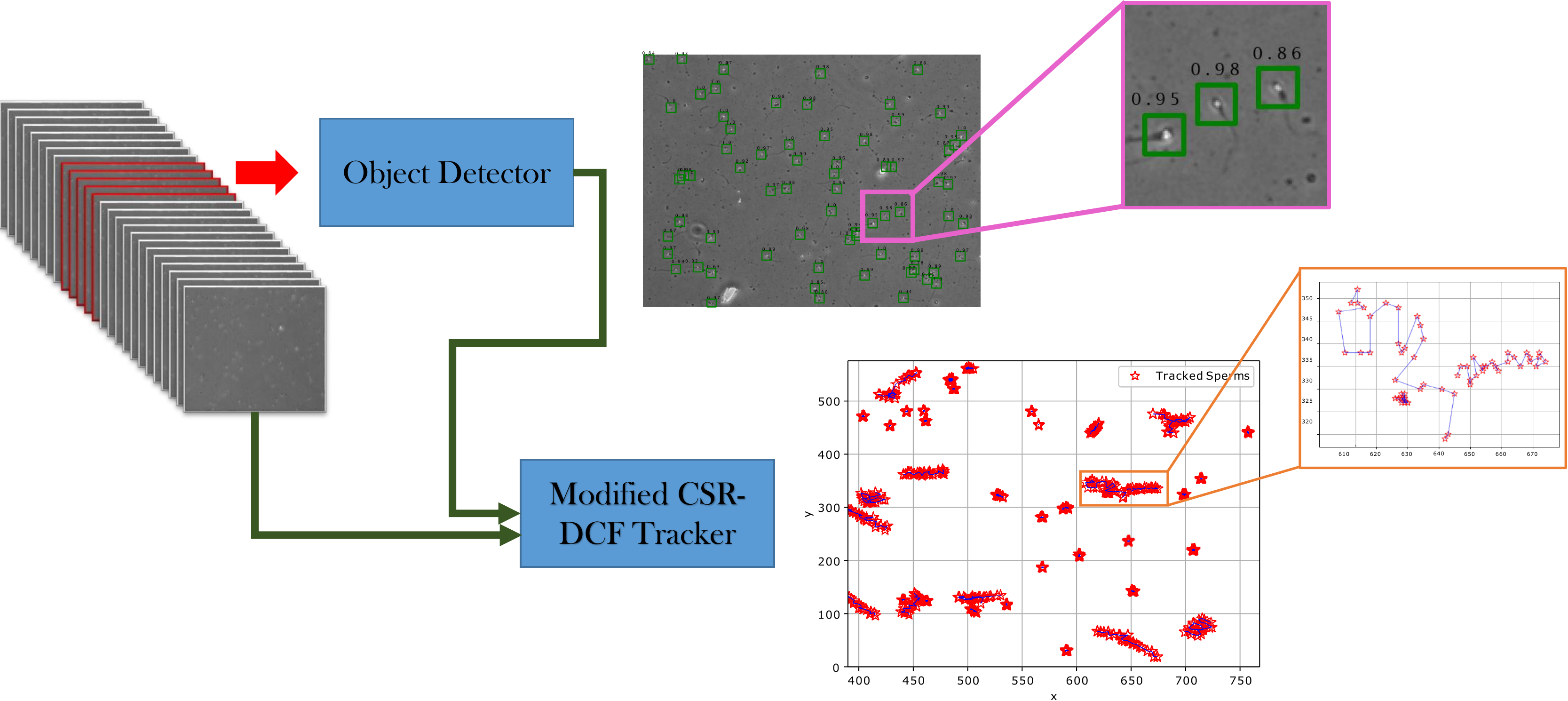}
	\caption{Schematic view of the proposed approach. Some consecutive frames of video sample would be concatenated and fed to the object detector. Then, the detected sperms and the video sample would be the input of the modified CSR-DCF tracker, and the results would be a list of tracked sperms in the video.}
	\label{fig:fig1}
\end{figure*}

The proposed approach is shown in Fig. \ref{fig:fig1} that consists of two steps: sperm detection and sperm tracking.
In the following, these steps are discussed in detail.

\subsection{Detection}
\label{31}
The first stage of our work is detecting sperms in the frames of a video sample.
We have used RetinaNet \cite{lin2018focal}, which is a deep fully convolutional network.
As described, a deep object detector, like RetinaNet, firstly attempts to extract object features, then, based on those features, detects the objects.
Sperms are small objects with few attributes like brightness, the special shape of head and tail, and motility.
Specially, motility attribute is significant because there might be other particles in the semen that look like sperms head and could cause false detections.
Utilizing motility attribute, the network learns to better distinguish sperms with other particles.

\begin{figure*}[!t]
	\centering
	\includegraphics[width=\linewidth]{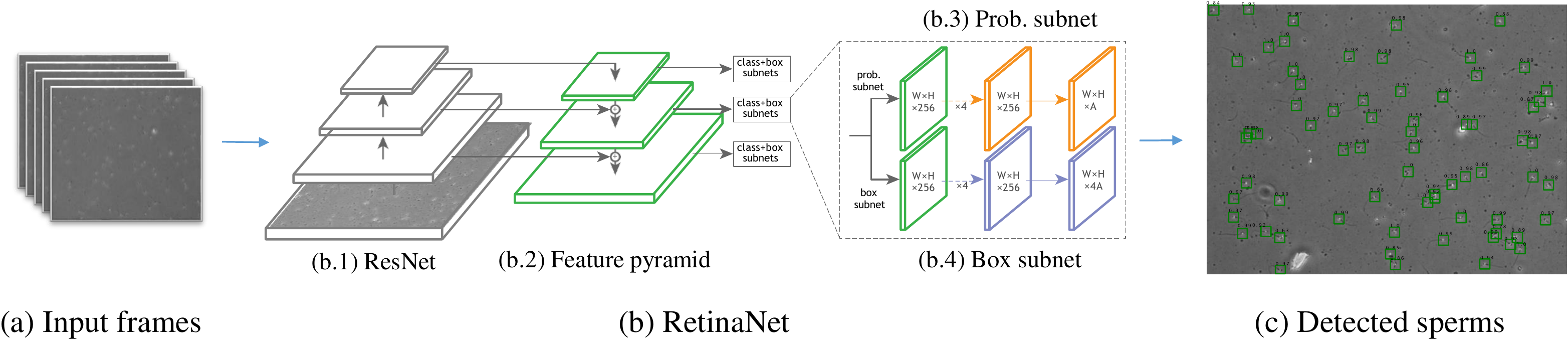}
	\caption{This schematic shows feeding the concatenation of five consecutive frames of a video for sperm detection.}
	\label{fig2}
\end{figure*}

If we feed one frame image to the object detector, the network cannot extract the motility attribute.
Our method is to feed the concatenation of consecutive frames of video samples to the network so that it can extract motility attributes.
The result reported in Section \ref{4} illustrates the superiority of our proposed method.
While training the network, we feed a concatenation of several frames and the ground-truth of the middle frame to the network.
Fig. \ref{fig2}, is an example of feeding a concatenation of five consecutive frames with the ground-truth of the middle frame to the network.
This way makes the network able to detect the sperm movement from previous and next frames and consider the motility attribute.

For testing the object detector, we concatenate each frame with its previous and next frames of the video.
It is noteworthy, that when the next frames or previous frames of the selected frame are not available (e.g., first or last frames of the video), we repeat the nearest frame instead of those unavailable frames.


As shown in Fig. \ref{fig2}, we have used RetinaNet as the base object detector.
RetinaNet is a deep convolutional neural network,  which consists of three main parts.
The first part is the backbone network, the second is a classifier, and the third part is used for box regressing \cite{lin2018focal}.
The backbone network in the RetinaNet that we used in this paper is the ResNet50 \cite{he2016deep}.
On the top of ResNet, feature pyramid network (FPN) \cite{lin2017feature} has been applied to improve feature extraction.
FPN is embedded in the deep convolutional neural networks to extract a multi-scale feature pyramid from the input image.
The classifier subnet is for detecting the possible spatial positions that the object could exist there.
The box regression subnet is to perform a regression from anchor boxes to ground-truth boxes \cite{lin2018focal}, and so after these two subnets, the object would be detected.

RetinaNet uses Focal Loss \cite{lin2018focal} as the loss functions, and this function has improved the performance of this object detector.
In the training stage, the loss of hard examples is more than easy examples, so Focal Loss focus on hard examples, by applying a modulating term to the cross-entropy loss function.\cite{lin2018focal}.
Applying this procedure has caused the Focal loss to improve the accuracy of RetinaNet \cite{lin2018focal}.
 
 

\subsection{Tracking}
\label{32}
The proposed method for tracking must be able to track the objects accurate and fast. The core of our introduced tracking algorithm is CSR-DCF \cite{lukezic_vojir_zajc_matas_kristan_2017}. CSR-DCF is a real-time and single-object tracker that works in semi-supervised mode. The CSR-DCF has achieved good tracking quality in the OTB100 \cite{wu2015object}, the VOT2015 \cite{kristan2015visual} and the VOT2016 \cite{Kristan2016a} benchmarks. One of our main reasons to use this method was because of these successes of this tracker algorithm and efficient tracking speed in processing.

In this research, due to the existence of multiple sperms in each frame, we need a multi-object tracker.
One of our novelties is modifying the CSR-DCF to a multi-object tracker.
We also improved it to be much more accurate in different conditions of noise, occlusion, the existence of false detections, and in high-density samples.
We call our proposed method, modified CSR-DCF.
It must be mentioned that our modified CSR-DCF algorithm, is not only based on the tracking but also is a combination of detected sperm in the frames and tracking algorithm.
Therefore, detection plays an important role in tracking algorithm, and as mentioned before, more accurate detection results in more accurate tracking.

The process of tracking firstly starts by initializing the CSR-DCF tracker \cite{lukezic_vojir_zajc_matas_kristan_2017} on the detected sperms in the first frame.
For each sperm in the first frame, the tracker would be initialized, then tracks it from the first frame to the second frame.
After that, the tracker would be initialized on the next sperm in the first frame and tracks it to the second frame, and so, this process continues for all the sperms in the first frame until all of them being tracked on the second frame.
At the next step, the tracker starts assigning the tracked sperms in the second frame to the detected sperms in the second frame.
If the tracked sperm and the detected sperm on the same frame have a distance of fewer than 15 pixels, they could be assigned.
The number 15 has been chosen because of our experimental experiences.
Each tracked sperm would be compared to all the detected sperms with a distance smaller or equal to 15 pixels on the same frame and would be assigned to the one detected sperm, in which their distance is the minimum.
Now, if there be more than one tracked sperm that should be assigned to one detected sperm, then the tracked sperm with less distance with that detected sperm would be assigned to it, and the previously tracked sperm would be rejected.
The rejected tracked sperm would be again compared to all the detected sperms because it may be assigned to another detected sperm, and if it could not be assigned then, it would be completely rejected.
All the tracked sperms that could not be assigned to any detected sperms would be considered as false tracks and would be rejected.
By doing this process, the possible wrong tracks would be removed.
If there is a detected sperm that, no tracked sperm is assigned to, would be considered as a new input sperm and starts a new track sequence from that frame.

Then the assigned tracked sperms and new input ones on the second frame would be tracked on the next frame, the same as the last procedure.
This process continues until all the sperms are tracked on the last frame of the video, which in our dataset is frame 25.
The described process makes the modified CSR-DCF a multi-object tracker.

\subsubsection{Missing tracks joiner}
\label{321}
Finally, to improve the accuracy of the tracking algorithm, and fixing separated tracks and the possible tracked sperms that are mistakenly rejected in different frames, we implement an algorithm called 'Missing Tracks joiner'.
This algorithm consists of four parts.
Firstly, it checks the sperm tracks that have not been started in the first frame, e.g., we call one of them, track A.
This algorithm compares track A with all other tracks that were abandoned in the previous frame that track A was started in it.
In the comparing process, if the number of sperms in two tracks be more than 3, the average distances traveled by both tracks is computed.
If the average distances are within a proper range of fewer than ten pixels per frame different, the function calculates the distance between the abandoned sperm in the previous frame and the started sperm in the current frame.
If the distance is smaller or equal to the max movement of the sperms between two consecutive frames in both tracks, plus ten (pixels), then the two tracks could be joined together.

In another condition, if the number of sperms of one track is less than 3, the algorithm only calculates the max movement of the sperms between two consecutive frames in both tracks. If the distance between the abandoned sperm and the started sperm be smaller or equal to the calculated value plus ten (pixels), then the two tracks could be joined.

\begin{figure}[!b]
	\centering
	\includegraphics[scale=0.625]{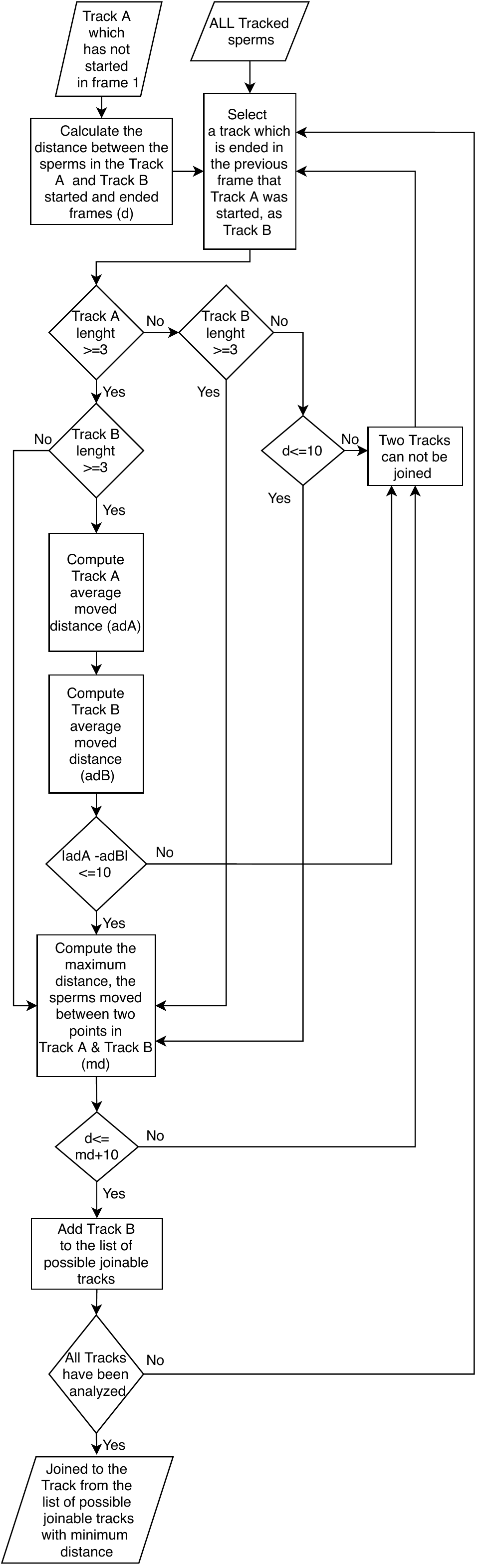}
	\caption{This flowchart shows the function of the first phase of the Missing Tracks Joiner algorithm.}
	\label{fig:joiner}
\end{figure}

In another case, if both tracks have less than three sperms, the algorithm calculates the distance between abandoned sperm and started sperm in the next frame.
If the calculated distance is smaller or equal to 10 pixels (the average movement of sperms is about 5 pixels per frame), the two tracks could be joined together.
It is important to say that the algorithm compares the started track, which we named track A with all other tracks ended in the previous frame that track A was started in it.
The most suitable track with all the described conditions would be joined to track A.
The flowchart of the first part of the Missing Tracks Joiner function is shown in Fig. \ref{fig:joiner}

The existence of false positives in the detected sperms can cause tracking failure and starting false tracks.
The false detected sperms could interrupt a running track and cause problems.
For example, consider a false detection that exists in frame x very close to a true positive sperm.
It may be possible that mistakenly the tracked sperm in the frame x would be assigned to the false detection instead of the true positive sperm.
Because of that, the true positive sperm may start a new track from frame x.
Now we would have two tracks, one started in frame x and one ended in frame x.
Although these two tracks are from one sperm, and the started track is the resume of the ended track, but they could not be joined because one is ended, and the other one is started in frame x.
This problem has been caused by false detection in frame x.
In the second phase, we aim to minimize this problem as much as possible.
In this phase, the function checks the sperm tracks that are left in a frame, and the tracks that are started in the same frame.
If the distance between the started sperm and abandoned sperm is smaller or equal to 10 pixels, it removes the last sperm location of the abandoned track and then joins it to the started track in the same frame.
Performing this part reduces the problems caused by false detections.

In the third part, the function attempts to solve the tracking failures that may be caused by false negatives (non-detected sperms). The abandoned tracks would be compared to the tracks started in two to five next frames, respectively.
The conditions for joining two tracks are like what has been described in the first phase of the Missing tracks joiner algorithm.
The only difference is that for the tracks with more than three sperms, the acceptable distance between the started sperm and the abandoned sperm has to be smaller or equal to the maximum movement of the sperms of two tracks per frame, plus 5, multiplied by the absent frames between two tracks.
If one of the two tracks have less than three sperms, the distance between abandoned and started tracks must be smaller or equal to the maximum movement of the sperms between two tracks, plus 5, multiplied by the absent frames.
For the tracks with less than three sperms, the acceptable distance between started sperm and abandoned sperm of two tracks is 10 pixels multiplied by the absent frames between two tracks.

Alternatively, in the fourth phase of the algorithm, according to the experience, if one sperm leaves the frame, it may re-enter in the next frames later.
In the fourth phase, abandoned sperm tracks around the border of a frame would be compared to the sperm tracks that were started around the border in the next five frames.
As mentioned, the average movements of sperms are around 5 pixels per frame.
With this information, the algorithm computes the distance between the started frame of one track and the abandoned frame of the other track, and if this distance is smaller or equal to 5 pixels multiplied by the absent frames between two tracks, two tracks would be joined.

\begin{figure}[!b]
	\centering
	\includegraphics[scale=0.625]{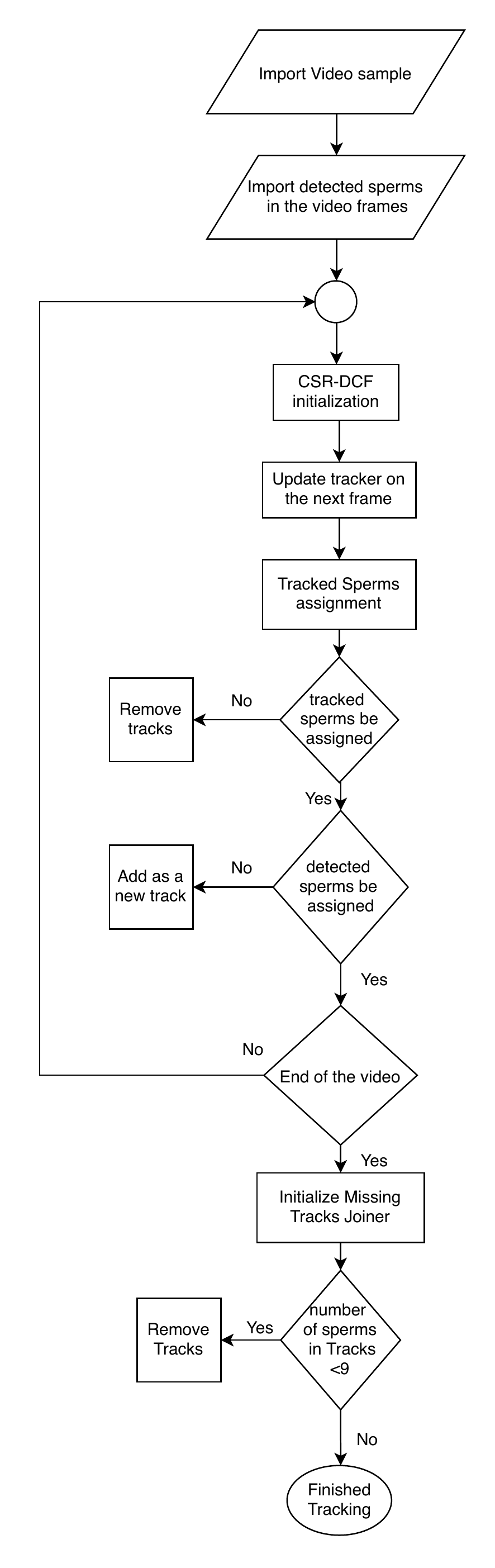}
	\caption{General flowchart of the modified CSR-DCF}
	\label{fig5}
\end{figure}

At last, based on our experience and have a dataset of 25 frames per video, the tracks with less than nine sperms would be removed. This is performed to remove the started tracks that are caused by false detected sperms in different frames.
With the applied algorithms, we tried to reduce false tracking as much as possible, and the results presented in Section \ref{4} illustrate improvement in sperm tracking accuracy.
The flowchart of the modified CSR-DCF algorithm is depicted in Fig. \ref{fig5}.

\section{Results and Comparison}
\label{4}
\subsection{Evaluation Methods}
\label{41}

In this section, we first describe the dataset used in our experiments.
Then, the evaluation methods are presented.
Finally, the results are reported and compared with the previous works.

\subsection{Dataset}
\label{sec:dataset}

We received our dataset from the author of \cite{arasteh2018multi} and applied tiny modifications to it. The dataset contains 36 different videos that were recorded in the Royan institute Research Lab. at Tehran.
The recorded videos are 8bit grayscale with 50 frames per second frame rate and $768\times576$ pixels resolution.
Each video consists of 25 frames.
As reported, each pixel in the video frames is 0.833 $\mu$m.
The number of sperms in the videos is in the range of 4 to 95 in each video, and in total, all the videos contain 1628 sperms.
The videos also differ in the value of noise and brightness and sperms collision.
This dataset has been manually labeled by the human experts, so we have the ground-truth for both detection and tracking phase.
We used One-fourth of the dataset for testing and the rest for the training.

\subsubsection{Detection Evaluation}
\label{411}

We have used the sperms manually annotated bounding boxes for evaluating the detected sperms.
As mentioned, we allocated one-fourth of the dataset for evaluating detection results.
The output of the neural network is detected boxes around the sperm cells, which could be correct or wrong.
The assignment of detected sperms to the annotations has been performed by applying the Intersection over Union (IoU) measure.
IoU score between the detected sperm and the annotation is calculated by computing the area of overlap and the area of union between them and then dividing the area of overlap by the area of union.
If the IoU score is more than a specific value, which we choose 0.5, detected sperm would be assigned to the annotation.
By this method, we have evaluated our detection method, and the results are reported in Section \ref{42}.

To measure the performance of the proposed approach, we use the Average precision (AP) metric.
AP is one of the most common evaluation metrics for object detectors with the following equation:

\begin{equation}
AP = \frac{\sum_{i=1}^{D}\left \{ Precision(i)\times Recall(i) \right \}}{Number\ of\ annotations}\label{eq:5} 
\end{equation}

where Precision and Recall formulas are: 

\begin{equation}
Recall=\frac{TP}{TP+FN}\label{eq:1}
\end{equation}

\begin{equation}
Precision=\frac{TP}{TP+FP}\label{eq:3}
\end{equation}

In these equations, \(TP\) (true positive) is the number of correct detected sperms, \(FP\) (false positive) is the number of the detected sperms that are not correct, and \(FN\) (false negative) is the number of the ground-truth sperms that have not been detected.
In Eq. \ref{eq:5}, \(D\) is the number of detected sperms that sorted by scores.

In addition to the \(AP\) metric, we also report Recall, Precision, \(F_1\)-measure, and Accuracy for the score threshold of 0.5.

\begin{equation}
Accuracy=\frac{TP}{TP+FP+FN}\label{eq:2}
\end{equation}

\begin{equation}
F_1=2\times \frac{Precision\times Recall}{Precision+Recall}\label{eq:4}
\end{equation}



\subsubsection{Tracking Evaluation}
\label{412}

To evaluate the sperm tracking algorithm, we use the manually annotated ground-truth of the sperm tracks.
The implemented evaluation method in this paper is inspired from the evaluation method used in \cite{arasteh2018multi}.
To compare one estimated track and one ground-truth track, if the non-overlapping frames are less than 6, we interpolate the missed frames (using bilinear interpolation and extrapolation).

\begin{figure}[!t]

	\centering
	\includegraphics[scale=0.6]{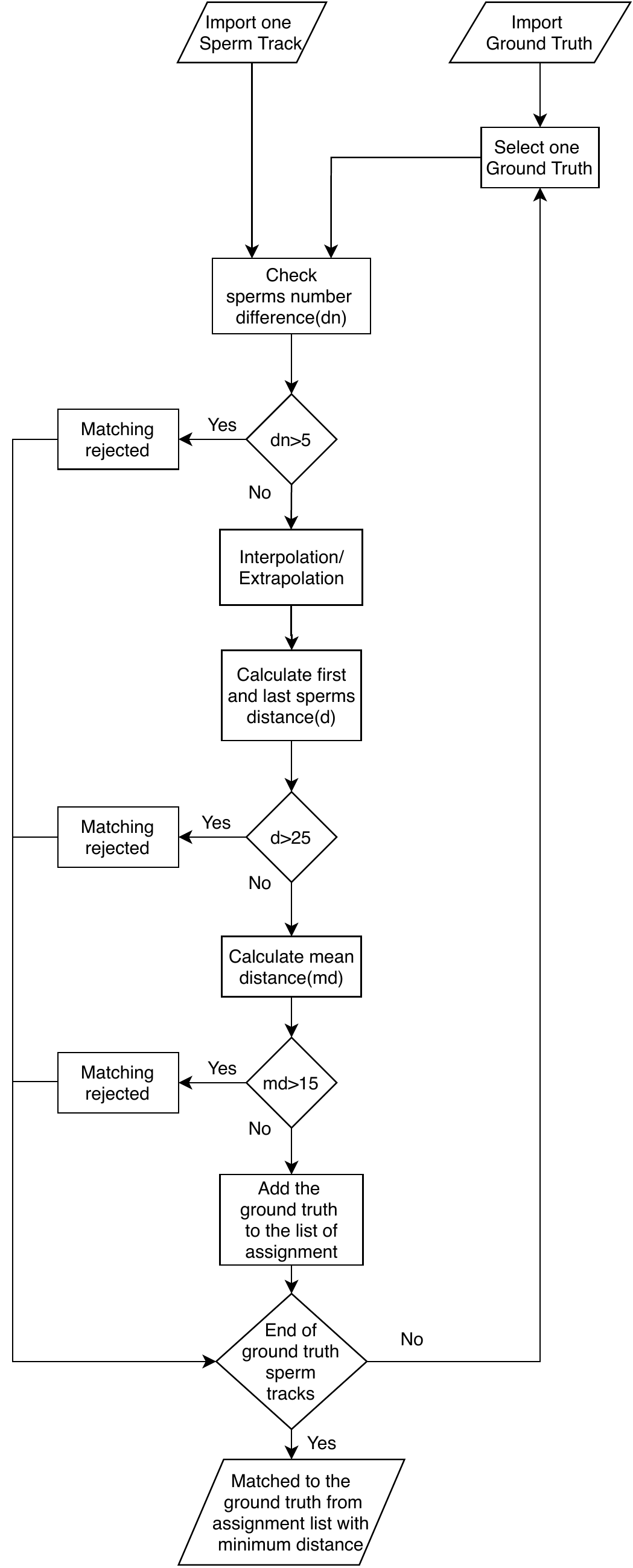}
	\caption{Flowchart of the evaluation method for one track}
	\label{fig3}

\end{figure}

After equalizing, we compare the equalized track with the ground-truth.
In this paper, we assumed that until five difference in the number of sperms in the track and the ground-truth is acceptable, and the average movement of the sperms are around 5 pixels per frame.
Because of this, for comparing the track and the ground-truth, if the distance between the first points and the last points be more than 25 pixels, they could not be matched.
And if the mean distance between the points of the track and the ground-truth be smaller or equal to 15 pixels, they may be possible to be matched.
Among all the ground-truth sperms, the function assigns the track to the ground-truth, with minimum mean distance.
The flowchart of the evaluation process has been depicted in Fig. \ref{fig3}.


\subsection{Results}
\label{42}
\subsubsection{Detection Results}
\label{421}

\begin{figure}[!b]
\centering
\includegraphics[scale=0.58]{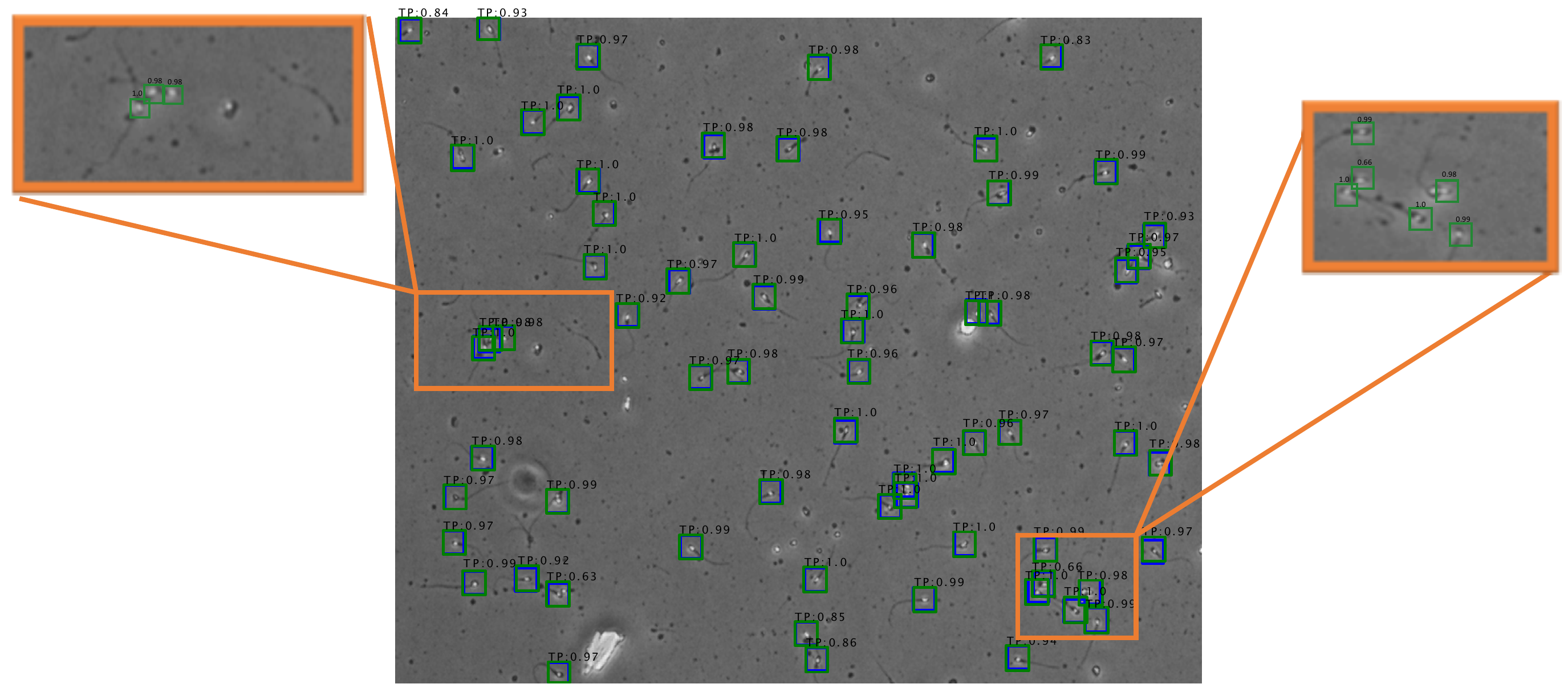}
\includegraphics[scale=0.41]{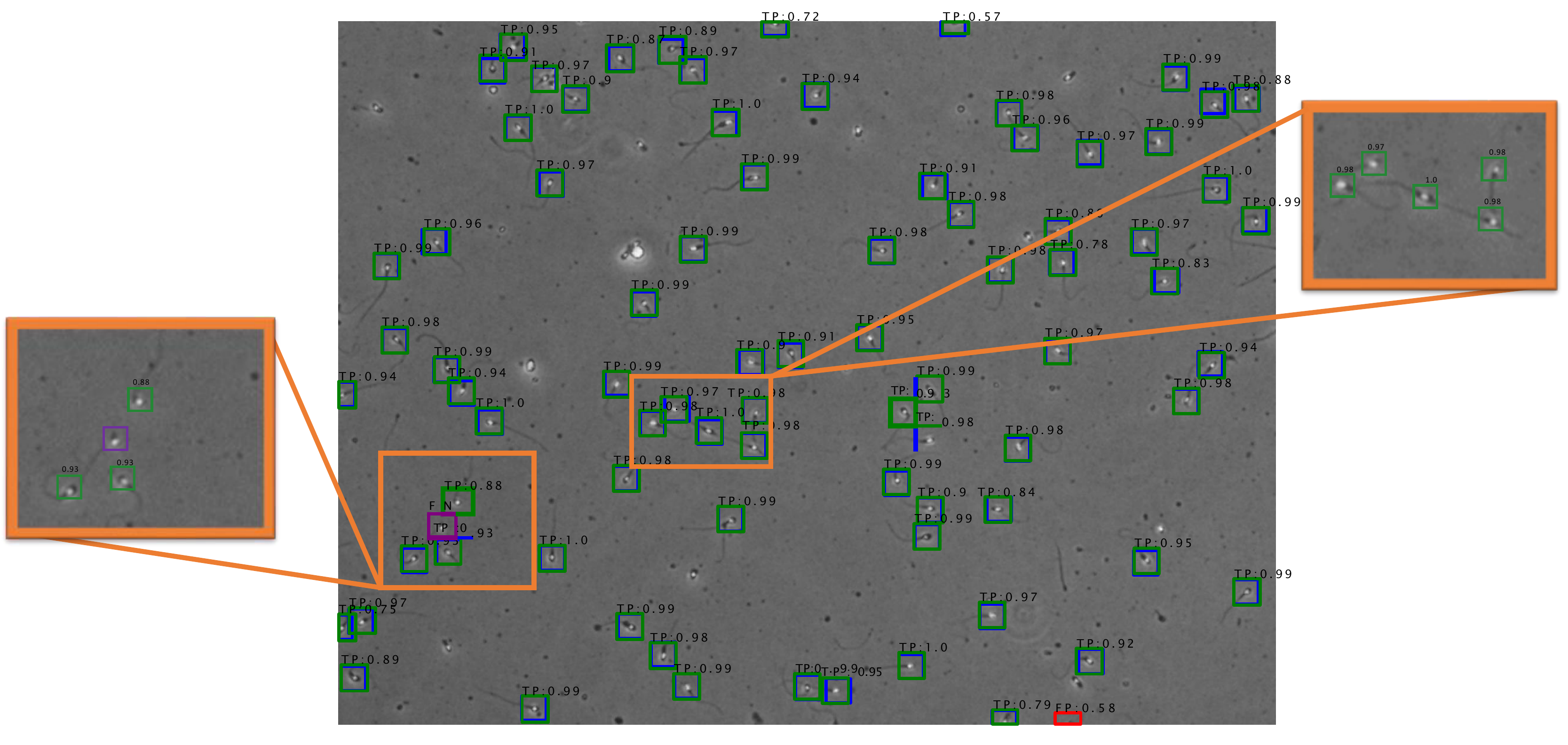}
\caption{Sample results of the proposed sperm detection algorithm}
\label{fig6}
\end{figure}

Some results of the proposed sperm detection algorithm is plotted in Fig. \ref{fig6}.
In this figure, the green boxes show true positives and the blue boxes are the corresponding ground-truth sperms.
The purple boxes are the false negative sperms, and the red boxes are the false positives.
The score of each of the detected sperms is written above the corresponding box.

\begin{figure*}[!t]

\centering
\subfloat[]{\label{main:aaa}\includegraphics[width=0.47\linewidth]{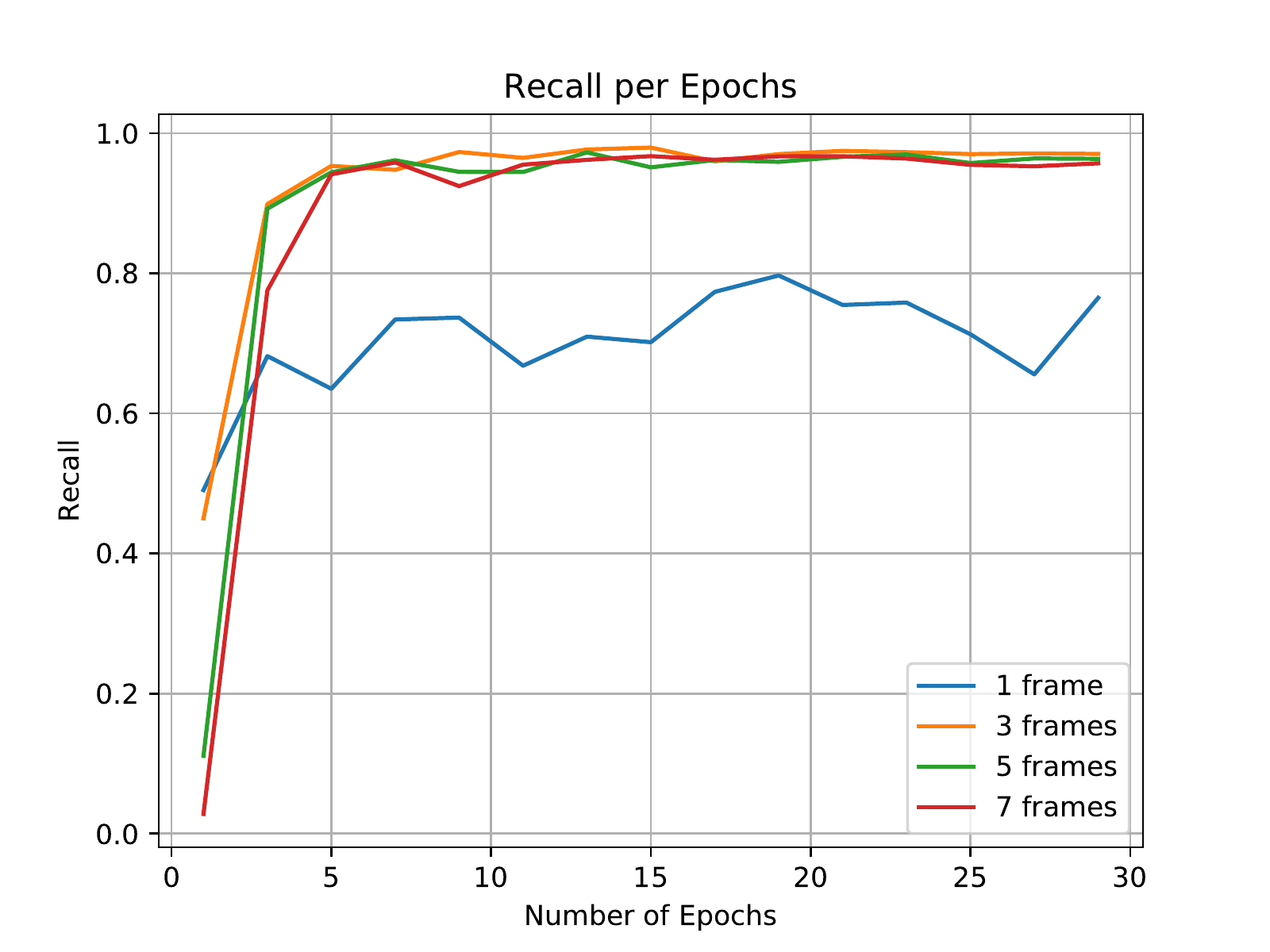}}
\subfloat[]{\label{main:bbb}\includegraphics[width=0.47\linewidth]{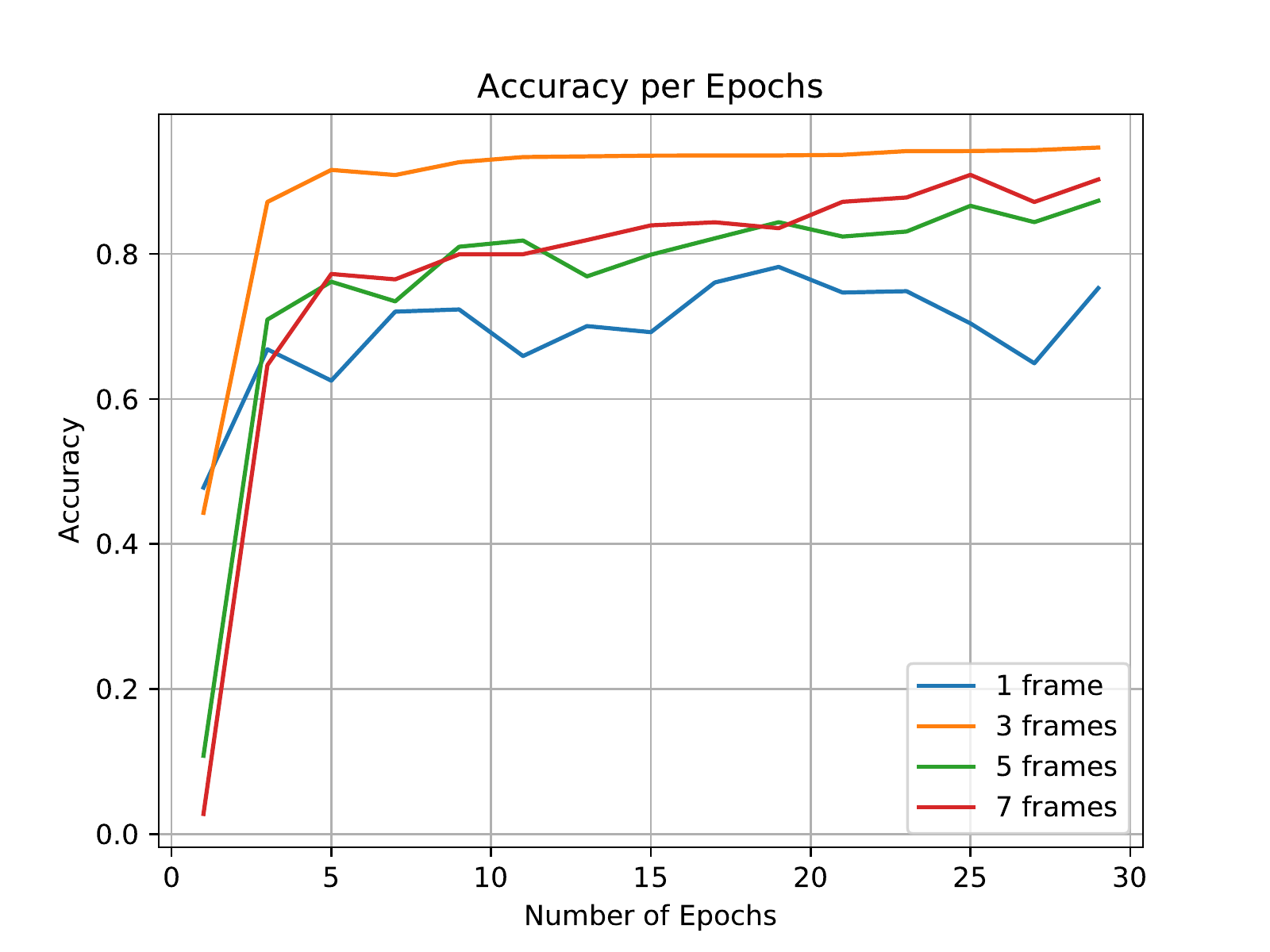}}
\newline	
\subfloat[]{\label{main:ccc}\includegraphics[width=0.47\linewidth]{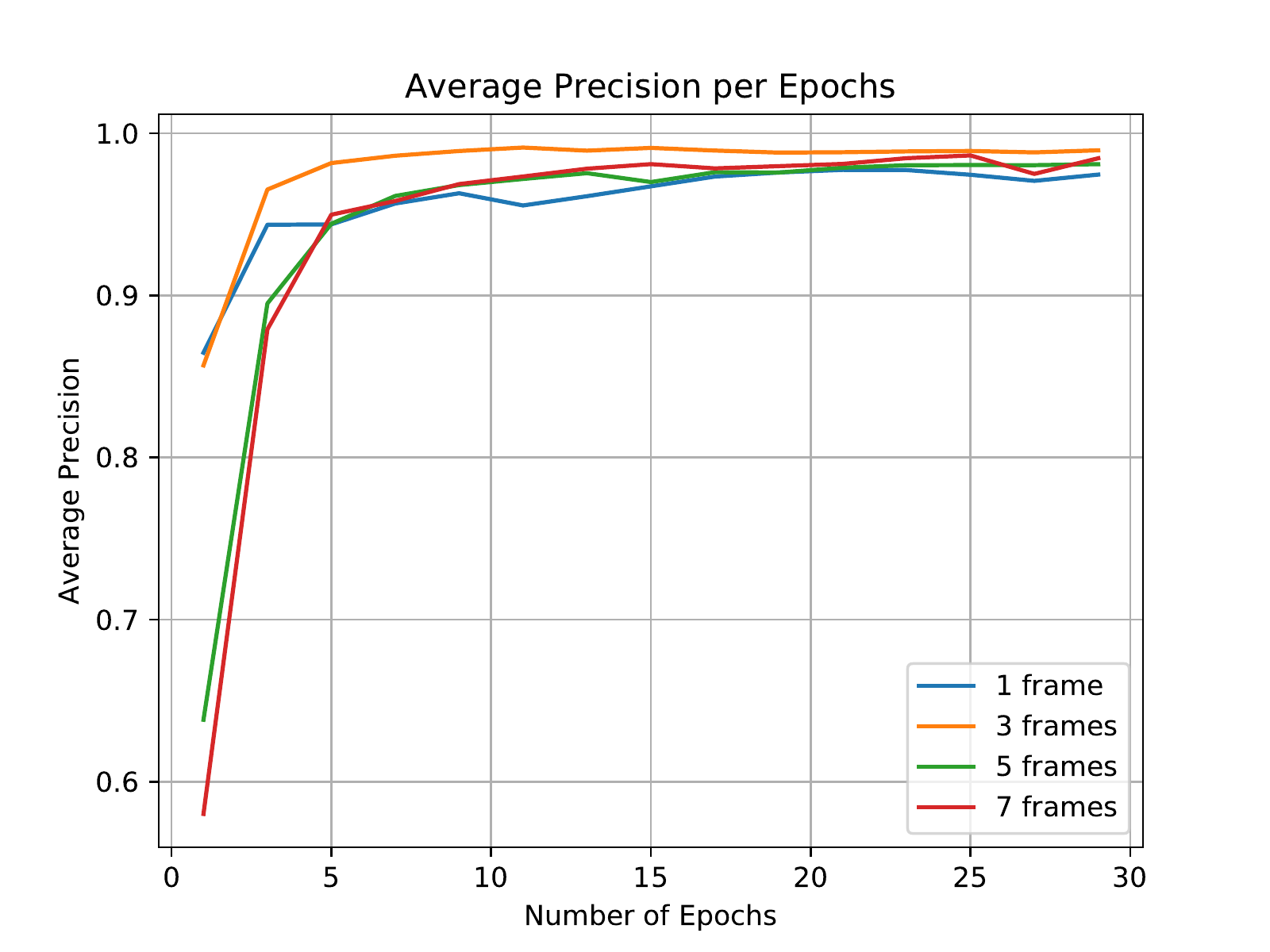}}
\subfloat[]{\label{main:ddd}\includegraphics[width=0.47\linewidth]{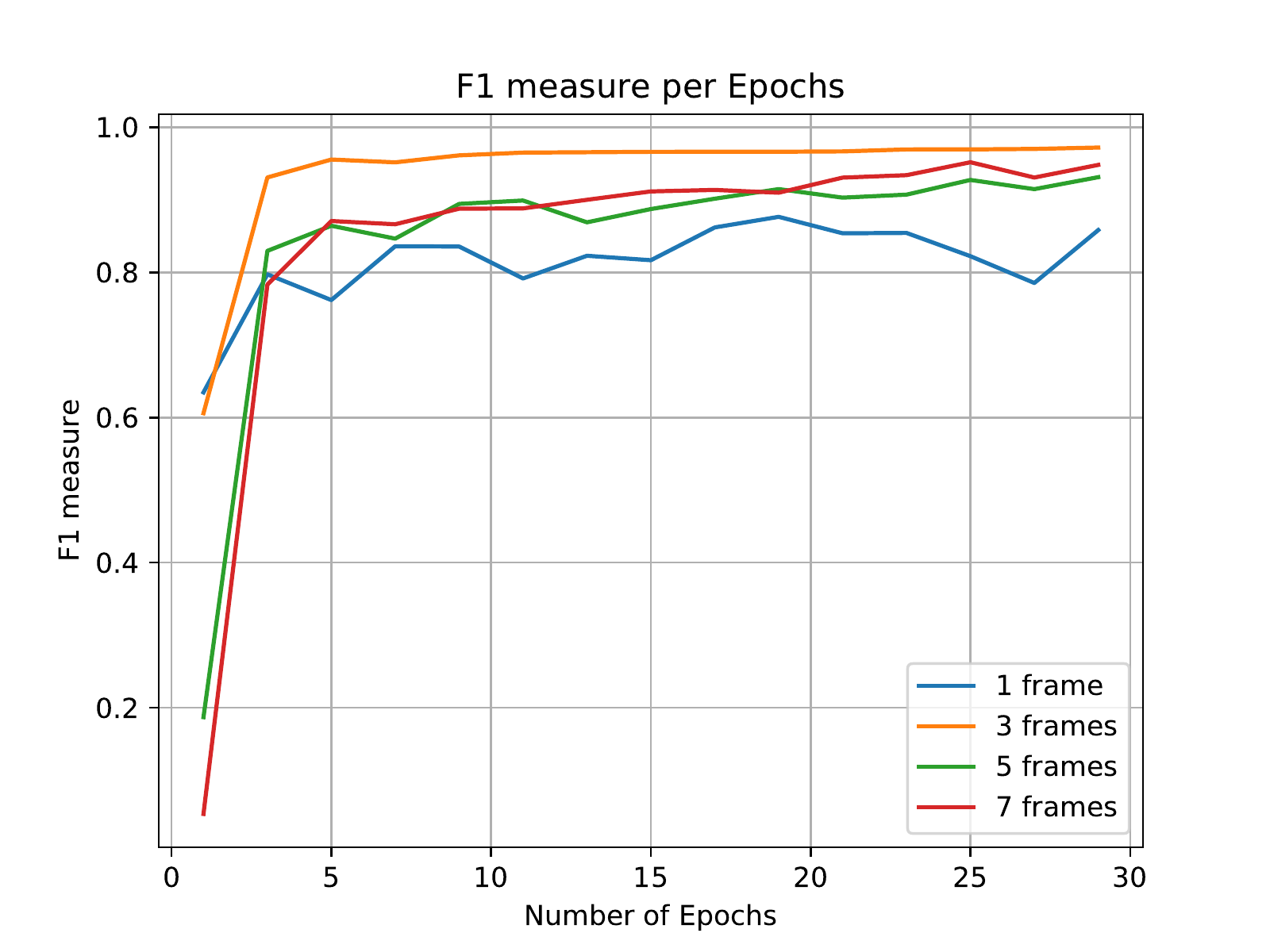}}

\caption{The different evaluation metrics for concatenation of the different number of frames.}
\label{fig9}
\end{figure*}

As stated, in this paper, we introduced a new method for detecting moving objects in frame sequences of a video.
This method works by feeding the concatenation of several consecutive frames to the network instead of only one frame.
We have evaluated our method by training the network with 3, 5, and 7 concatenated consecutive frames and compared the results with using only one frame.
The comparison results are represented in Fig. \ref{fig9}.
We trained each concatenation, three different times with 675 steps, which is the number of our training dataset, until 29 epochs. Then we reported the averaged metrics values between 3 different times of training.
It is noteworthy that data augmentation methods like rotation, translation, and flipping, have been used for training. 

It is obvious from Fig. \ref{fig9} that concatenation of consecutive frames results in much better training output.
This method can make a good leap in detecting moving objects in the video frames, and also can be used in many other domains.
It can be understood from Fig. \ref{fig9} that, among different concatenated frames, concatenation of 3 consecutive frames delivers the most accurate results.
Based on the obtained results, we resumed our training model based on the concatenation of 3 frames. 

We resumed our training based on concatenation of 3 frames until 40 epochs with 10000 steps and by applying data augmentation methods like rotation, translation, and flipping.
The best-reached detection results, have been implemented on the tracking algorithm and are presented in Table \ref{table1}.
These results have been tested on one-fourth of the dataset, which contains nine different videos.
Unfortunately, the dataset that many papers in the domain of sperm detection have used is not available, and many other papers just have reported the tracking results.

The system that we used in the detection phase was Tesla T4 GPU with 12 GB RAM, which was provided by the \href{https://colab.research.google.com/}{Google Colaboratory Notebooks}. We implemented the neural network with Keras\cite{chollet2015keras}.
In the tracking phase, we implemented our codes using the OpenCV library\cite{2015opencv}  on a laptop with Intel Core i7-9750H CPU and 16 GB RAM. 

\begin{table*}[!t]
\centering
\begin{tabular}{|c|c|c|c|c|}
\hline
\begin{tabular}[c]{@{}c@{}}AP\end{tabular} & Recall & Accuracy & Precision & \begin{tabular}[c]{@{}c@{}}F1\end{tabular} \\ \hline
99.1                                                         & 98.7  & 96.3    & 97.4     & 98.1                                                \\ \hline
\end{tabular}
\caption{Sperm detection results using the concatenation of 3 consecutive frames.}
\label{table1}
\end{table*}


\subsubsection{Tracking Results}
\label{422}

The modified CSR-DCF algorithm is dependent on the detected sperm in the video frames.
At the first experiment, we gave our algorithm, the ground-truth instead of the detected sperms by the network.
This means that we are testing our algorithm, consuming that the detection accuracy is 100\%.
The result was 100\% in all metrics with no false positives or false negatives.
This illustrates that, how much the detection results be more accurate, then the tracking result also will be more precise to even 100\%.

At the next experiment, we used our detected sperms with the reported accuracy and other metrics.
Our tracking algorithm has been tested on 9 videos containing 411 sperms.
The result of the tracker algorithm at these experiments are presented in Table \ref{table2}.
In this table, we have reported the tracking algorithm results for the 9 videos.
The videos include different intensity and sperm numbers.
After that, in the last row of Table \ref{table2} we have reported the result of testing the tracker algorithm on all of the 9 videos.

\begin{table*}[!t]
\centering
\caption{Modified CSR-DCF results using ground-truth and detected boxes}
\begin{tabular}{|l|l|l|l|l|l|l|l|l|l|} 
\hline
Sample~                                                        & Number of & \multicolumn{4}{c|}{Ground-Truth Boxes} &  \multicolumn{4}{c|}{Detection}      \\ 
\cline{3-10}
No.                                                            & sperms    & Recall               & Precision                      & Accuracy                  & F1  & Recall                      & Precision              & Accuracy                     & F1   \\ 
\hline
1                                                            & 12        & 100                 & 100                            & 100                       & 100       & 91.66                        & 100                   & 91.66                          & 95.64         \\ \hline
2                                                            & 14        & 100                  & 100                            & 100                       & 100        & 100                         & 100                 & 100                        & 100       \\ \hline
3                                                            & 22        & 100                  & 100                            & 100                       & 100        & 100                         & 100                    & 100                          & 100         \\ \hline
4                                                            & 26        & 100                  & 100                            & 100                       & 100        & 96.15                        & 96.15                   & 92.59                         & 96.15        \\ \hline
5                                                            & 32        & 100                  & 100                            & 100                       & 100        & 100                         & 100                 & 100                        & 100       \\ \hline
6                                                            & 49        & 100                  & 100                            & 100                       & 100        & 97.95                        & 96.00                   & 94.11                         & 96.96        \\ \hline
7                                                            & 66        & 100                  & 100                            & 100                       & 100        & 95.45                        & 96.92                   & 92.64                         & 96.17        \\ \hline
8                                                            & 95        & 100                  & 100                            & 100                       & 100        & 96.84                        & 93.87                   & 91.08                         & 9533          \\ \hline
9                                                            & 95        & 100                  & 100                            & 100                       & 100        & 97.89                        & 93.93                  & 92.07                        & 95.86       \\ \hline
All & 411       & 100                  & 100                            & 100                       & 100        & 97.32                       & 95.92                  & 93.45                        & 96.61      \\
\hline
\end{tabular}
\label{table2}
\end{table*}

\begin{figure}[!b]

\centering
\label{main:a}\includegraphics[width=0.75\linewidth]{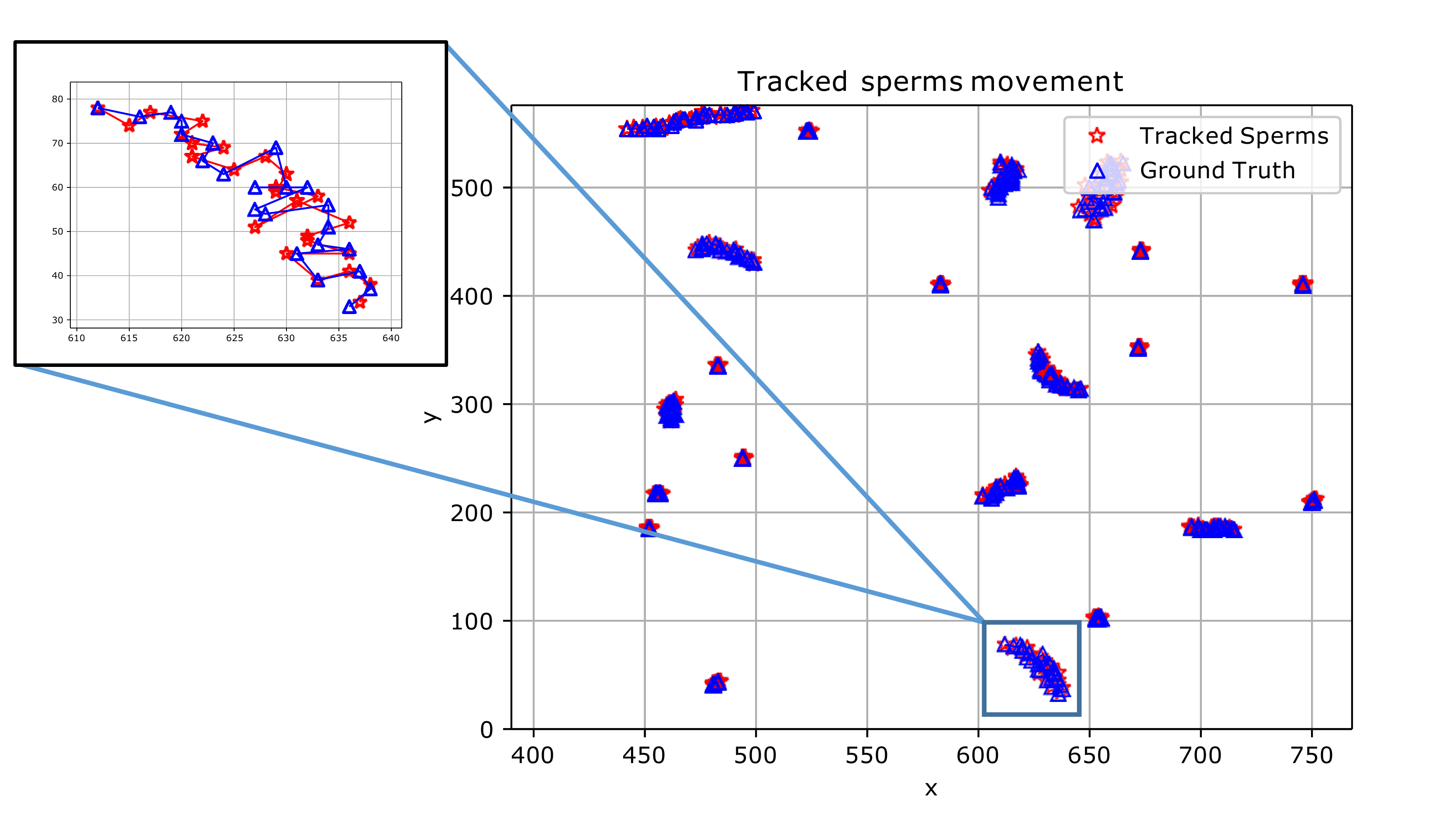}
\centering
\label{main:b}\includegraphics[width=0.75\linewidth]{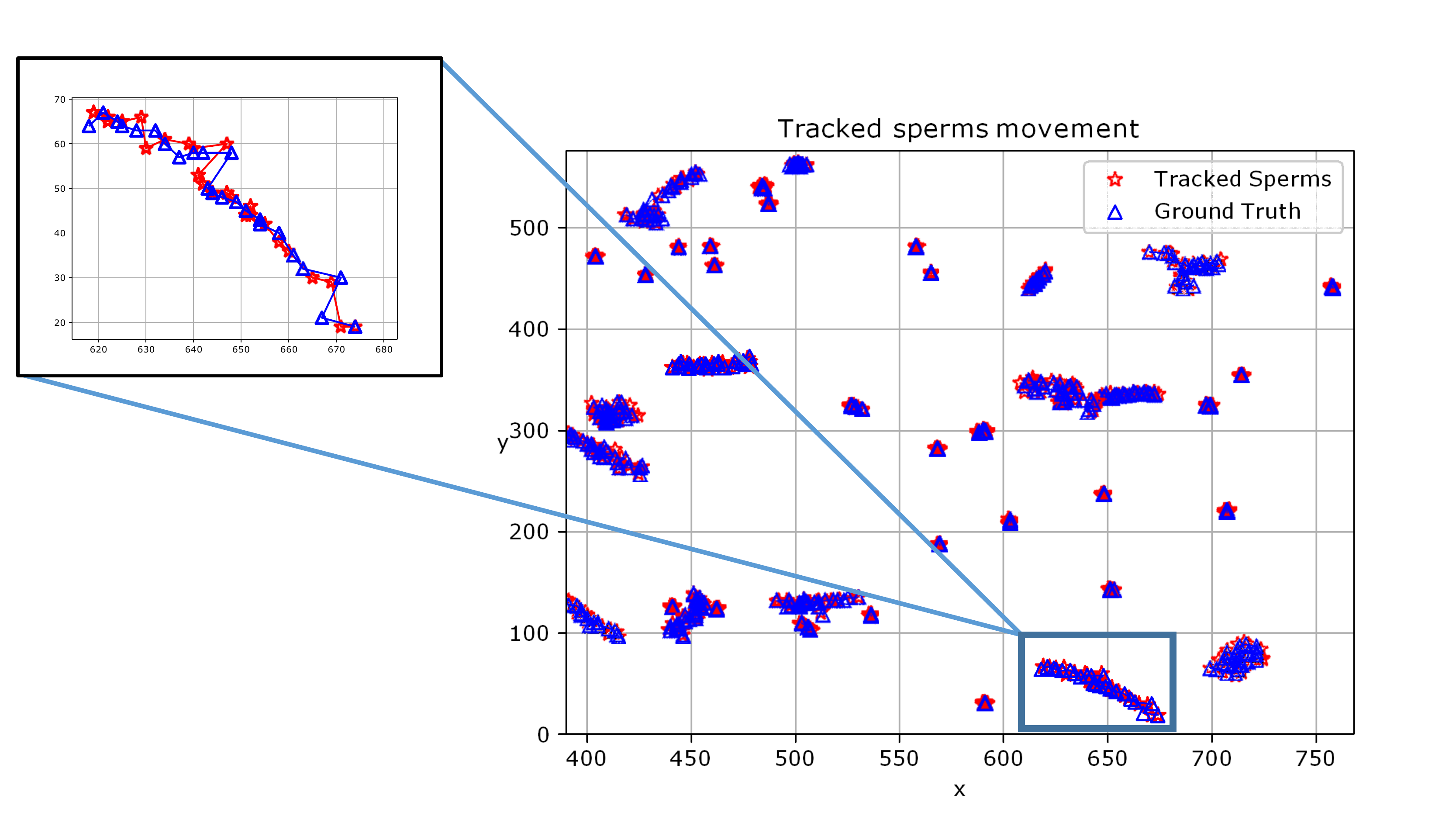}

\caption{In these two images, we can see the tracked sperms and the ground-truth sperm tracks that are shown by red stars and blue triangles, respectively. Some sperms are immotile, and some of them move at different speeds in different directions.}
\label{fig7}
\end{figure}

Some tracked sperms with ground-truth have been depicted in Fig. \ref{fig7}.

\subsubsection{Motility Parameters Results}
\label{423}

In this section, we extract some motility parameters from the detection and tracking results. For every tracked sperm in each video sample, we measured velocity straight line (VSL), velocity curvilinear (VCL), velocity average pathway (VAP), the straightness (STR), and the linearity (LIN).

The mentioned parameters are defined as, VSL: the straight line distance between the first and the last points of the tracked sperm divided by time elapsed ($\mu$m/s), VCL: the average speed of the tracked sperm, which is calculated by dividing the sum of all the distances a sperm travels from a point to another point, by the time elapsed ($\mu$m/s), VAP: the smoothed version of VCL, it is calculated by dividing the sum of all the distances a sperm passes along it's smoothed average path by the time elapsed ($\mu$m/s), STR: it is the ratio of VSL/VAP (\%), and LIN is the ratio of VSL/VCL (\%).
Based on these parameters, we clustered the sperms into six categories: Immotile sperms, slow sperms, sperms with medium velocity, rapid sperms, non-progressive, and progressive sperms.

The six categories are defined as rapid (VAP > MVV), sperms with medium velocity (LVV < VAP < MVV), slow(VAP < LVV), progressive (VAP > MVV and STR > standardized threshold of STR.), non-progressive (Sperms which are not progressive, nor immotile) and the immotile sperms are ones that do not move while tracking.\cite{Kathiravan2011}
According to \cite{Rijsselaere2003}, We defined our parameters like: Medium VAP cut-off (MVV)=50$\mu$m/s, Low VAP cut-off(LVV)=30$\mu$m/s, and the standardized threshold of STR=70\%. 

Since each pixel in our video samples is 0.833 $\mu$m, and because the bounding boxes around the detected immotile sperms could vary a little between different frames, we defined the immotile sperms those that, their VAP be less than 10 pixels/s or 8.33 $\mu$m/s. In table \ref{table4}, five motility parameters have been reported, and in table \ref{table5}, the nine validation video samples have been clustered into six categories.

\begin{table*}
\centering
\caption{The mean of 5 motility parameters for the 9 validation video samples. Each parameter value is the mean of that parameter between all of the tracked sperms in the video sample.}
\begin{tabular}{|l|l|l|l|l|l|l|} 
\hline
\begin{tabular}[c]{@{}l@{}}Sample\\No.\end{tabular} & \begin{tabular}[c]{@{}l@{}}Number of \\detected \\sperms\end{tabular} & VSL($\mu$m/s)    & VAP($\mu$m/s)     & VCL($\mu$m/s)     & STR(\%)    & LIN(\%)     \\ 
\hline
1                                                   & 11                                                                    & 120.15 & 134.35 & 219.56 & 87.48 & 60.60  \\ 
\hline
2                                                   & 14                                                                    & 53.87  & 59.18  & 99.88  & 82.80 & 42.87  \\ 
\hline
3                                                   & 22                                                                    & 86.81  & 102.20 & 192.75 & 84.67 & 52.52  \\ 
\hline
4                                                   & 26                                                                    & 41.66  & 51.42  & 78.34  & 76.26 & 35.79  \\ 
\hline
5                                                   & 32                                                                    & 51.51  & 60.84  & 83.10  & 73.96 & 44.41  \\ 
\hline
6                                                   & 50                                                                    & 33.28  & 40.82  & 65.30  & 65.24 & 34.82  \\ 
\hline
7                                                   & 65                                                                    & 38.75  & 57.15  & 109.36 & 72.35 & 29.16  \\ 
\hline
8                                                   & 98                                                                    & 31.38  & 42.82  & 73.90  & 70.32 & 29.06  \\ 
\hline
9                                                   & 99                                                                    & 24.96  & 39.03  & 72.09  & 56.18 & 22.48  \\
\hline
\end{tabular}
\label{table4}
\end{table*}

\begin{table*}
\centering
\caption{Clustering the video samples into 6 categories}
\begin{tabular}{|l|l|l|l|l|l|l|l|} 
\hline
\begin{tabular}[c]{@{}l@{}} Sample\\~No. \end{tabular} & \begin{tabular}[c]{@{}l@{}}Number of \\detected \\sperms \end{tabular} & Immotile      & Slow          & Medium      & Rapid         & Non-Progressive & Progressive    \\ 
\hline
1                                                      & 11                                                                     & 0 (0.0\%)     & 0 (0.0\%)     & 0 (0.0\%)   & 11 (100.0\%)  & 2 (18.18\%)     & 9 (81.81\%)    \\ 
\hline
2                                                      & 14                                                                     & 5 (35.71\%)   & 0 (0.0\%)     & 0 (0.0\%)   & 9 (64.28\%)   & 1 (7.14\%)      & 8 (57.14\%)    \\ 
\hline
3                                                      & 22                                                                     & 0 (0.0\%)     & 0 (0.0\%)     & 0 (0.0\%)   & 22 (100.0\%)  & 5 (22.72\%)~    & 17 (77.27\%)~  \\ 
\hline
4                                                      & 26                                                                     & 14 (53.84\%)  & 3 (11.53\%)   & 0 (0.0\%)   & 9 (34.61\%)   & 5 (19.23\%)~    & 7 (26.92\%)~   \\ 
\hline
5                                                      & 32                                                                     & 12 (37.50\%)  & 0 (0.0\%)     & 3 (9.37\%)  & 17 (53.12\%)  & 7 (21.87\%)~    & 13 40.62\%)~   \\ 
\hline
6                                                      & 50                                                                     & 22 (44.0\%)   & 4 (8.0\%)     & 6 (12.0\%)~ & 18 (36.0\%)~  & 14 (28.00\%)~   & 14 (28.00\%)~  \\ 
\hline
7                                                      & 65                                                                     & 21 (32.30\%)~ & 13 (20.0\%)~  & 1 (1.53\%)~ & 30 (46.15\%)~ & 24 (36.92\%)~   & 20 30.76\%)~   \\ 
\hline
8                                                      & 98                                                                     & 47 (47.95\%)~ & 15 (15.30\%)~ & 3 (3.06\%)~ & 33 (33.67\%)~ & 29 (29.59\%)~   & 22 (22.44\%)~  \\ 
\hline
9                                                      & 99                                                                     & 51 (51.51\%)  & 13 (13.13\%)~ & 4 (4.04\%)~ & 31 (31.31\%)~ & 33 (33.33\%)~   & 15 (15.15\%)~  \\ 
\hline
All                                                    & 417                                                                    & 172           & 48            & 17          & 180           & 120             & 125            \\
\hline
\end{tabular}
\label{table5}
\end{table*}

\subsection{Comparison}
\label{43}
In table \ref{table3}, we have compared the results of our proposed tracking algorithm with CSR-DCF \cite{lukezic_vojir_zajc_matas_kristan_2017}, and Hybrid Dynamic Bayesian Network algorithm (HDBN) \cite{arasteh2018multi}.
Based on table \ref{table3}, the superiority of our tracking algorithm is clear.

\begin{table*}[!t]
\centering
\caption{Comparison between our proposed method and other methods}
\begin{tabular}{|l|l|l|l|l|} 
\hline
Method  & Recall         & Precision      & F1             & Accuracy        \\ 
\hline
CSR-DCF & 90.51          & 89.63          & 90.06          & 81.93           \\ 
\hline
HDBN    & 64.14          & 95.66          & 76.79          & -               \\ 
\hline
Ours    & \textbf{97.32} & \textbf{95.92} & \textbf{96.61} & \textbf{93.45}  \\
\hline
\end{tabular}
\label{table3}
\end{table*}

\section{Conclusion}
\label{5}
In this paper, we aim to detect and track sperms in phase-contrast microscopy image sequences.
The proposed approach operates in two stages.
In the first stage, we introduce a new method for detecting moving objects in a video.
Based on this method, for training and testing, instead of feeding one frame to the network, a concatenation of the frame with its previous and next frames, would be fed to the network.
This proposed approach helps the object detector to be able to extract motility attributes, too.
In our paper, we used RetinaNet, a deep fully convolutional neural network, as the object detector.
As we know, sperms are motile objects, so by applying the introduced, we observed remarkable improvement in the output results of the network.
The final obtained F1 score for the detection stage is 98.1\%. 

In the next stage, we introduced a new multi-object tracker, which is called modified CSR-DCF.
This algorithm is a detection-based tracking, and, its accuracy is very high.
The central core of our proposed tracker is the CSR-DCF algorithm.
We modified the CSR-DCF algorithm and added some other functions like missing tracks joiner to it, so it became a very accurate multi-object tracker.
It performs very well, even in the existence of noise, sperms colliding, occlusion, and false detection.
We obtained 96.61\% F1 score from evaluation of our proposed tracker method.



\section{Code Availability}
\label{6}
We have made our code available on (\url{https://github.com/mr7495/Sperm_detection_and_tracking}).
\section*{Acknowledgment}
We thank Royan Institute for providing raw sperm data and Dr.Abollah Arasteh for sharing this dataset and the ground-truth to us.
We also thank \href{https://github.com/fizyr}{Fizyr}
, whose implemented RetinaNet with Keras  \cite{chollet2015keras} on GitHub, and the \href{https://colab.research.google.com/}{Google Colab server} for providing free and powerful GPU.

\bibliographystyle{abbrv} 

\bibliography{arxiv}

\begin{thebibliography}{10}

\bibitem{amann2004andrology}
R.~P. Amann and D.~F. Katz.
\newblock {Andrology lab corner*: Reflections on casa after 25 years}.
\newblock {\em Journal of andrology}, 25(3):317--325, 2004.

\bibitem{arasteh2018multi}
A.~Arasteh, B.~V. Vahdat, and R.~S. Yazdi.
\newblock {Multi-Target Tracking of Human Spermatozoa in Phase-Contrast
  Microscopy Image Sequences using a Hybrid Dynamic Bayesian Network}.
\newblock {\em Scientific reports}, 8(1):5068, 2018.

\bibitem{bermant2019deep}
P.~C. Bermant, M.~M. Bronstein, R.~J. Wood, S.~Gero, and D.~F. Gruber.
\newblock Deep machine learning techniques for the detection and classification
  of sperm whale bioacoustics.
\newblock {\em Scientific reports}, 9(1):1--10, 2019.

\bibitem{beya2015automatic}
O.~Beya, M.~Hittawe, D.~Sidib{\'{e}}, and F.~Meriaudeau.
\newblock {Automatic detection and tracking of animal sperm cells in microscopy
  images}.
\newblock In {\em 2015 11th International Conference on Signal-Image Technology
  {\&} Internet-Based Systems (SITIS)}, pages 155--159. IEEE, 2015.

\bibitem{chollet2015keras}
F.~Chollet and Others.
\newblock keras, 2015.

\bibitem{el2003human}
A.~A. El-Ghobashy and C.~R. West.
\newblock {The human sperm head: a key for successful fertilization}.
\newblock {\em Journal of andrology}, 24(2):232--238, 2003.

\bibitem{groenewald1991preprocessing}
A.~Groenewald and E.~Botha.
\newblock Preprocessing and tracking algorithms for automatic sperm analysis.
\newblock In {\em COMSIG 1991 Proceedings: South African Symposium on
  Communications and Signal Processing}, pages 64--68. IEEE, 1991.

\bibitem{Gussi2005}
I.~Gussi.
\newblock {Clinical Gynecologic Endocrinology and Infertility}.
\newblock {\em Acta Endocrinologica (Bucharest)}, 2005.

\bibitem{he2016deep}
K.~He, X.~Zhang, S.~Ren, and J.~Sun.
\newblock {Deep residual learning for image recognition}.
\newblock In {\em Proceedings of the IEEE conference on computer vision and
  pattern recognition}, pages 770--778, 2016.

\bibitem{helmstaedter2013connectomic}
M.~Helmstaedter, K.~L. Briggman, S.~C. Turaga, V.~Jain, H.~S. Seung, and
  W.~Denk.
\newblock {Connectomic reconstruction of the inner plexiform layer in the mouse
  retina}.
\newblock {\em Nature}, 500(7461):168, 2013.

\bibitem{hesar2018multiple}
H.~D. Hesar, H.~A. Moghaddam, A.~Safari, and P.~Eftekhari-Yazdi.
\newblock {Multiple sperm tracking in microscopic videos using modified GM-PHD
  filter}.
\newblock {\em Machine Vision and Applications}, 29(3):433--451, 2018.

\bibitem{hicks2019machine}
S.~A. Hicks, J.~M. Andersen, O.~Witczak, V.~Thambawita, P.~Halvorsen, H.~L.
  Hammer, T.~B. Haugen, and M.~A. Riegler.
\newblock {Machine learning-based analysis of sperm videos and participant data
  for male fertility prediction}.
\newblock {\em Scientific reports}, 9(1):1--10, 2019.

\bibitem{javadi2019novel}
S.~Javadi and S.~A. Mirroshandel.
\newblock {A novel deep learning method for automatic assessment of human sperm
  images}.
\newblock {\em Computers in biology and medicine}, 109:182--194, 2019.

\bibitem{Kathiravan2011}
P.~Kathiravan, J.~Kalatharan, G.~Karthikeya, K.~Rengarajan, and G.~Kadirvel.
\newblock {Objective Sperm Motion Analysis to Assess Dairy Bull Fertility Using
  Computer-Aided System - A Review}, 2011.

\bibitem{katz1985real}
D.~F. Katz, R.~O. Davis, B.~A. Delandmeter, and J.~W. Overstreet.
\newblock {Real-time analysis of sperm motion using automatic video image
  digitization}.
\newblock {\em Computer methods and programs in biomedicine}, 21(3):173--182,
  1985.

\bibitem{kheirkhah2018modified}
F.~M. Kheirkhah, H.~R.~S. Mohammadi, and A.~Shahverdi.
\newblock Modified histogram-based segmentation and adaptive distance tracking
  of sperm cells image sequences.
\newblock {\em Computer methods and programs in biomedicine}, 154:173--182,
  2018.

\bibitem{Kheirkhah2019}
F.~M. Kheirkhah, H.~R.~S. Mohammadi, and A.~Shahverdi.
\newblock {Efficient and robust segmentation and tracking of sperm cells in
  microscopic image sequences}.
\newblock {\em IET Computer Vision}, 2019.

\bibitem{Kristan2016a}
M.~Kristan, A.~Leonardis, J.~Matas, M.~Felsberg, R.~Pflugfelder, L.~{\v{C}}.
  Zajc, T.~Vojir, G.~H{\"{a}}ger, A.~Luke{\v{z}}i{\v{c}}, and G.~Fernandez.
\newblock {The Visual Object Tracking VOT2016 challenge results}.
\newblock Springer, oct 2016.

\bibitem{kristan2015visual}
M.~Kristan, J.~Matas, A.~Leonardis, M.~Felsberg, L.~Cehovin, G.~Fernandez,
  T.~Vojir, G.~Hager, G.~Nebehay, and R.~Pflugfelder.
\newblock {The visual object tracking vot2015 challenge results}.
\newblock In {\em Proceedings of the IEEE international conference on computer
  vision workshops}, pages 1--23, 2015.

\bibitem{lecun2015deep}
Y.~LeCun, Y.~Bengio, and G.~Hinton.
\newblock {Deep learning}.
\newblock {\em nature}, 521(7553):436, 2015.

\bibitem{lin2017feature}
T.-Y. Lin, P.~Doll{\'{a}}r, R.~Girshick, K.~He, B.~Hariharan, and S.~Belongie.
\newblock {Feature pyramid networks for object detection}.
\newblock In {\em Proceedings of the IEEE conference on computer vision and
  pattern recognition}, pages 2117--2125, 2017.

\bibitem{lin2018focal}
T.-Y. Lin, P.~Goyal, R.~Girshick, K.~He, and P.~Doll{\'{a}}r.
\newblock {Focal loss for dense object detection}.
\newblock {\em IEEE Transactions on Pattern Analysis and Machine Intelligence},
  2018.

\bibitem{liu2016ssd}
W.~Liu, D.~Anguelov, D.~Erhan, C.~Szegedy, S.~Reed, C.-Y. Fu, and A.~C. Berg.
\newblock {Ssd: Single shot multibox detector}.
\newblock In {\em European conference on computer vision}, pages 21--37.
  Springer, 2016.

\bibitem{lukezic_vojir_zajc_matas_kristan_2017}
A.~Lukezic, T.~Vojir, L.~C. Zajc, J.~Matas, and M.~Kristan.
\newblock {Discriminative Correlation Filter with Channel and Spatial
  Reliability}.
\newblock {\em 2017 IEEE Conference on Computer Vision and Pattern Recognition
  (CVPR)}, 2017.

\bibitem{ma2015deep}
J.~Ma, R.~P. Sheridan, A.~Liaw, G.~E. Dahl, and V.~Svetnik.
\newblock {Deep neural nets as a method for quantitative structure--activity
  relationships}.
\newblock {\em Journal of chemical information and modeling}, 55(2):263--274,
  2015.

\bibitem{menkveld2001semen}
R.~Menkveld, W.~Y. Wong, C.~J. Lombard, A.~M.~M. Wetzels, C.~M.~G. Thomas,
  H.~M. W.~M. Merkus, and R.~P.~M. Steegers-Theunissen.
\newblock {Semen parameters, including WHO and strict criteria morphology, in a
  fertile and subfertile population: an effort towards standardization of
  in-vivo thresholds}.
\newblock {\em Human Reproduction}, 16(6):1165--1171, 2001.

\bibitem{mortimer2015future}
S.~T. Mortimer, G.~van~der Horst, and D.~Mortimer.
\newblock {The future of computer-aided sperm analysis}.
\newblock {\em Asian journal of andrology}, 17(4):545, 2015.

\bibitem{2015opencv}
OpenCV.
\newblock Open source computer vision library, 2015.

\bibitem{oquab2014learning}
M.~Oquab, L.~Bottou, I.~Laptev, and J.~Sivic.
\newblock Learning and transferring mid-level image representations using
  convolutional neural networks.
\newblock In {\em Proceedings of the IEEE conference on computer vision and
  pattern recognition}, pages 1717--1724, 2014.

\bibitem{world1999laboratory}
W.~H. Organisation.
\newblock {\em {WHO laboratory manual for the examination of human semen and
  sperm-cervical mucus interaction}}.
\newblock Cambridge university press, 1999.

\bibitem{redmon2017yolo9000}
J.~Redmon and A.~Farhadi.
\newblock {YOLO9000: better, faster, stronger}.
\newblock In {\em Proceedings of the IEEE conference on computer vision and
  pattern recognition}, pages 7263--7271, 2017.

\bibitem{ren2015faster}
S.~Ren, K.~He, R.~Girshick, and J.~Sun.
\newblock {Faster r-cnn: Towards real-time object detection with region
  proposal networks}.
\newblock In {\em Advances in neural information processing systems}, pages
  91--99, 2015.

\bibitem{Rijsselaere2003}
T.~Rijsselaere, A.~{Van Soom}, D.~Maes, and A.~{De Kruif}.
\newblock {Effect of technical settings on canine semen motility parameters
  measured by the Hamilton-Thorne analyzer}.
\newblock {\em Theriogenology}, 2003.

\bibitem{riordon2019deep}
J.~Riordon, C.~McCallum, and D.~Sinton.
\newblock {Deep learning for the classification of human sperm}.
\newblock {\em Computers in Biology and Medicine}, page 103342, 2019.

\bibitem{schmidhuber2015deep}
J.~Schmidhuber.
\newblock Deep learning in neural networks: An overview.
\newblock {\em Neural networks}, 61:85--117, 2015.

\bibitem{shi2006computer}
L.~Z. Shi, J.~Nascimento, M.~W. Berns, and E.~L. Botvinick.
\newblock {Computer-based tracking of single sperm}.
\newblock {\em Journal of biomedical optics}, 11(5):54009, 2006.

\bibitem{sorensen2008multi}
L.~S{\o}rensen, J.~{\O}stergaard, P.~Johansen, and M.~de~Bruijne.
\newblock {Multi-object tracking of human spermatozoa}.
\newblock In {\em Medical Imaging 2008: Image Processing}, volume 6914, page
  69142C. International Society for Optics and Photonics, 2008.

\bibitem{urbano2016automatic}
L.~F. Urbano, P.~Masson, M.~VerMilyea, and M.~Kam.
\newblock {Automatic tracking and motility analysis of human sperm in
  time-lapse images}.
\newblock {\em IEEE transactions on medical imaging}, 36(3):792--801, 2016.

\bibitem{wu2015object}
Y.~Wu, J.~Lim, and M.-H. Yang.
\newblock {Object tracking benchmark}.
\newblock {\em IEEE Transactions on Pattern Analysis and Machine Intelligence},
  37(9):1834--1848, 2015.

\bibitem{xiong2015human}
H.~Y. Xiong, B.~Alipanahi, L.~J. Lee, H.~Bretschneider, D.~Merico, R.~K.~C.
  Yuen, Y.~Hua, S.~Gueroussov, H.~S. Najafabadi, T.~R. Hughes, and Others.
\newblock {The human splicing code reveals new insights into the genetic
  determinants of disease}.
\newblock {\em Science}, 347(6218):1254806, 2015.

\bibitem{zhang2018robotic}
Z.~Zhang, C.~Dai, J.~Huang, X.~Wang, J.~Liu, J.~Zhang, S.~Moskovtsev,
  C.~Librach, K.~Jarvi, and Y.~Sun.
\newblock {Robotic Immobilization of Motile Sperm}.
\newblock In {\em 2018 IEEE International Conference on Robotics and Automation
  (ICRA)}, pages 2676--2681. IEEE, 2018.

\bibitem{zinaman1996evaluation}
M.~J. ZINAMAN, M.~L. UHLER, E.~Vertuno, S.~G. FISHER, and E.~D. CLEGG.
\newblock {Evaluation of computer-assisted semen analysis (CASA) with IDENT
  stain to determine sperm concentration}.
\newblock {\em Journal of Andrology}, 17(3):288--292, 1996.

\end{thebibliography}

\end{document}